\documentclass{edm_article}

\usepackage[table]{xcolor}
\usepackage{url}
\usepackage{hyperref}
\usepackage{cleveref}
\usepackage{siunitx}
\usepackage{amsmath}
\usepackage{bm}
\begin{document}

\title{Investigating the Robustness of Knowledge Tracing Models in the Presence of Student Concept Drift}

\numberofauthors{7}
\author{
\alignauthor
Morgan Lee
\thanks{Denotes equal contribution}
\thanks{Correspondence to mplee@wpi.edu}\\
       \affaddr{Worcester Polytechnic Institute}\\
\alignauthor
Artem Frenk
$^*$\\
       \affaddr{Worcester Polytechnic Institute}\\
\alignauthor Eamon Worden\\
\affaddr{Worcester Polytechnic Institute}\\
\and 
\alignauthor Karish Gupta\\
\affaddr{Worcester Polytechnic Institute}\\
\alignauthor Thinh Pham\\
\affaddr{Worcester Polytechnic Institute}\\
\and 
\alignauthor Ethan Croteau\\
\affaddr{Worcester Polytechnic Institute}\\
\alignauthor Neil Heffernan\\
\affaddr{Worcester Polytechnic Institute}\\
}

\maketitle

\begin{abstract}
Knowledge Tracing (KT) has been an established problem in the educational data mining field for decades, and it is commonly assumed that the underlying learning process being modeled remains static. Given the ever-changing landscape of online learning platforms (OLPs), we investigate how concept drift and changing student populations can impact student behavior within an OLP through testing model performance both within a single academic year and across multiple academic years. Four well-studied KT models were applied to five academic years of data to assess how susceptible KT models are to concept drift. Through our analysis, we find that all four families of KT models can exhibit degraded performance, Bayesian Knowledge Tracing (BKT) remains the most stable KT model when applied to newer data, while more complex, attention based models lose predictive power significantly faster. To foster more longitudinal evaluations of KT models, the data used to conduct our analysis is available at \url{https://osf.io/hvfn9/?view_only=b936c63dfdae4b0b987a2f0d4038f72a}.
 \end{abstract}

\section{Introduction} \label{intro}

To extract meaning from large amounts of data, one may generally assume that the underlying processes which generate said data are static, or at least relatively stable over long periods of time. One of the core goals of educational data mining (EDM) is, of course, using data to model aspects of students' learning processes. These models can then be used to predict, explain, or challenge our current understanding of how students learn in online learning platforms (OLPs). Even in cases where the goal is to better suit individual learners' needs, the methodology remains the same: create a model of student behavior and learning by analyzing previously collected data, identify components of that model which may fail to account for individual differences, and update the model to account for the ways in which particular learners differ. These powerful methodologies have allowed researchers to detect the affective state of students \cite{botelho_improving_2017,calvo_affect_2010}, model the procedural acquisition of knowledge \cite{corbett_knowledge_1994}, predict student success \cite{kovacic_early_2010}, and detect problematic or unhelpful student behavior \cite{baker_detecting_2010, vanacore2024effect}. Many of these approaches have been practiced for multiple decades by now, meaning multiple generations of learners have had their learning processes studied, aggregated, and modeled. As it currently stands, the goal of providing high-quality, scalable educational software that responds to individual student needs \cite{shemshack_systematic_2020} is closer than ever before.

As the use of student modeling techniques becomes ever more ubiquitous, it is necessary to revisit the assumptions that guide our practice. We \textit{assume} that data collected from different learners represents the same underlying learning process. We \textit{assume} that the ways in which students learn that are measurable by scientists and practitioners remain consistent. Given the maturation of EDM as a field and the availability of learner data spanning generations of learners, perhaps it is now possible to verify that our assumptions are correct, or at least to identify the circumstances where they are safe assumptions to make.

The educational best-practices of 30 years ago are obviously not the educational best-practices of the modern day. Educational policy has shifted towards meticulous measurement of student progress \cite{Hallinger01032011}, identifying failing schools \cite{nicolaidou_understanding_2005}, and standardizing subject curricula to better facilitate rigorous measurement \cite{popkewitz_educational_2004}. Simultaneously, OLPs rose in popularity, automating student practice and proliferating student engagement data \cite{ritter_cognitive_2007,heffernan_assistments_2014}. These educational platforms have also matured since their creation, and every pedagogical and cosmetic change to these platforms could impact the way students interact with these platforms. Even ignoring educational policy changes, students are individuals in a large and changing world, and world events which change how humans relate to one another impact students as much as anyone else. In a particularly extreme example, an entire generation of students experienced learning losses due to the COVID-19 pandemic \cite{donnelly_learning_2022}. In a changing world, how can we be sure our modeling techniques are still valid? 

More specifically, we wish to understand how Knowledge Tracing (KT) models are impacted by changes in student populations over time. KT is a foundational problem of the educational data mining field, and as such we intend to investigate how a number of different modeling techniques behave when applied outside of their temporal context. To achieve this, we borrow from Data Mining literature the concept of dataset shift, and discuss its applicability to online learning platforms. We then propose a methodology for evaluating KT models both within their temporal context and across student populations, taking steps to ensure that our datasets contain similar exercise banks and Knowledge Concepts. We then apply this methodology to four well-studied KT models and examine how each model performs outside of its temporal context. We then conclude by discussing the implications of our findings, as well as limitations and different directions future work could take to overcome said limitations.

Our analysis was guided by the following research questions:
\begin{enumerate}
    \item[\textbf{RQ1.}] How robust are KT models to changing student populations?
    \item[\textbf{RQ2.}] Does the complexity of a KT model impact its susceptibility to concept drift?
\end{enumerate}
In addition to the listed analyses above, there is a notable lack of knowledge tracing datasets that span multiple years. Alongside this paper we will be releasing a publicly available dataset containing student interaction data spanning multiple academic years, allowing other researchers to evaluate their own models under controlled conditions of "aging."
\section{Background} \label{bg}
In this section, we introduce and discuss literature relevant to our investigation of KT model robustness. First, we introduce KT as a specific machine learning task and discuss specific models which will be investigated in this work (Section \ref{kt-models}). Next, we discuss frameworks for analyzing the drift of software systems from their original contexts (Section \ref{dist-shifts}). Finally, we discuss relevant prior work investigating the generalization of KT models (Section \ref{prior}).

\subsection{Knowledge Tracing} \label{kt-models}

Long established in EDM literature, KT is defined as a many-to-many time series binary classification problem attempting to predict the correctness of future student responses based on prior performance. Numerous machine learning architectures have been applied to this task, including Factorization Machines \cite{vie_knowledge_2019} and psychometric models like Item Response Theory \cite{yeung_deep-irt_2019}. Shen et al. \cite{shen_survey_2024} provides a comprehensive survey of historical and contemporary methods. In this work, we will be replicating four well-studied KT models: Bayesian Knowledge Tracing, Performance Factors Analysis, Deep Knowledge Tracing, and Self-Attentive Knowledge Tracing.

\paragraph{Bayesian Knowledge Tracing} \label{bkt}

Originally proposed by Corbett \& Anderson \cite{corbett_knowledge_1994} and deeply connected to mastery learning \cite{bloom_learning_1968}, Bayesian Knowledge Tracing (BKT) models the acquisition of knowledge as a latent Markov process. The student's knowledge state is modeled as a latent variable that is noisily observed through performance on exercises in an intelligent tutoring system. Exercises in said tutoring system are tagged with Knowledge Components (KCs) signifying related items, and items tagged with the same KC are treated as having uniform difficulty. Mastery of different KCs is computed independently, meaning that mastery of one KC is completely independent of other KCs. Due to the assumptions made about students' learning processes, each parameter of a BKT model has a direct interpretation that is explainable to teachers and other educational practitioners \cite{williamson_effects_2021}. The latent learning process was originally modeled as one-way, modeling students as unable to forget KCs once they have been mastered, but later model variants explore forgetting behavior \cite{qiu_does_2011}, individual estimations of prior knowledge \cite{yudelson_individualized_2013}, and contextual guess and slip parameters \cite{hutchison_more_2008}. 

\begin{figure}
    \centering
    \includegraphics[width=0.3\textwidth]{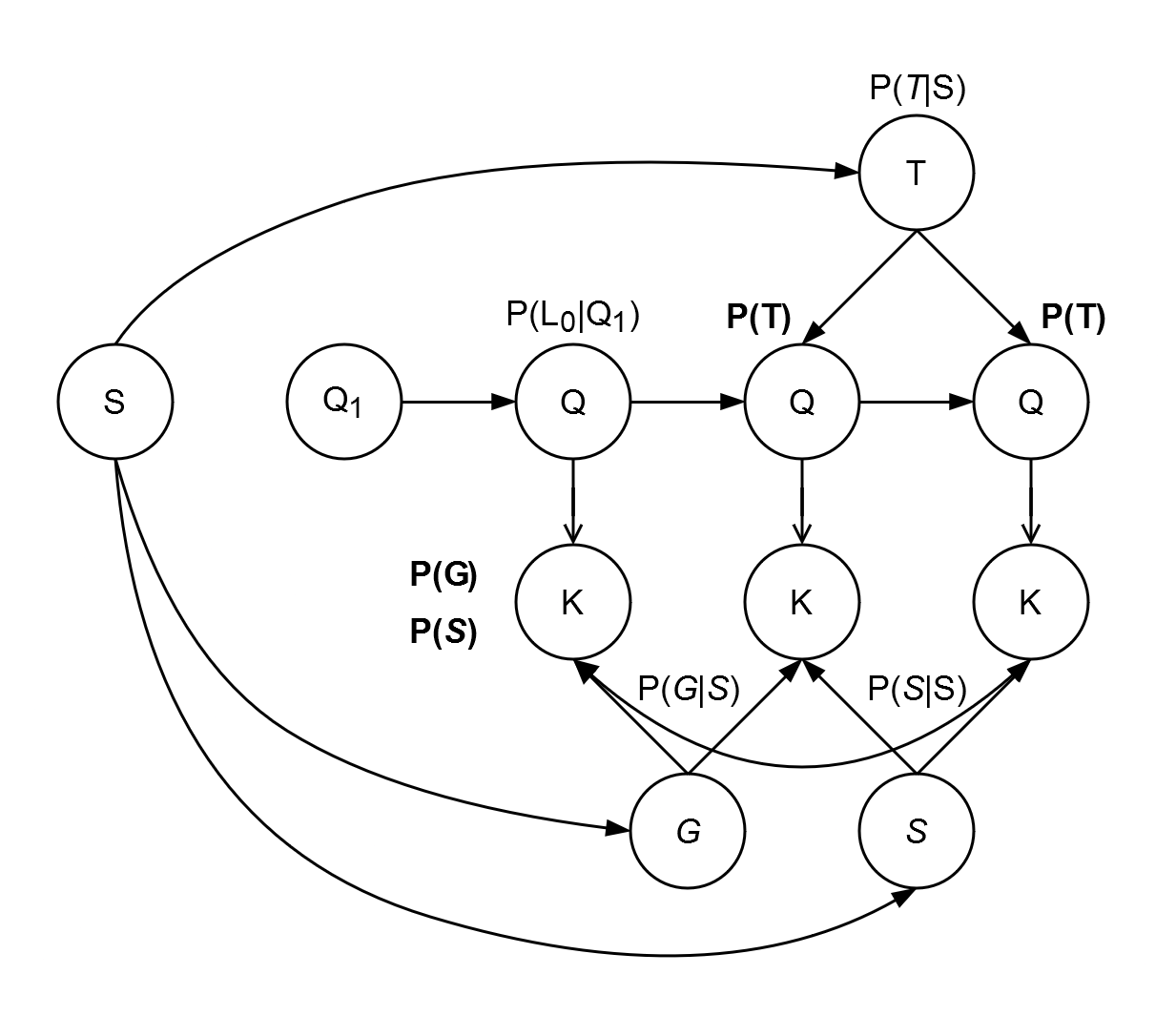}
    \Description[A network graph of BKT nodes]{Network graph where each node is a BKT model parameter. }
    \caption{Model architecture for BKT}
    \label{fig:BKT} 
\end{figure}

\paragraph{Performance Factors Analysis} \label{pfa}

Closely related to Learning Factors Analysis \cite{cen_learning_2006}, Performance Factors Analysis (PFA) was first proposed in Pavlik et al. \cite{pavlik_performance_2009} as a logistic regression based alternative to traditional Knowledge Tracing models. Rather than sequentially modeling a student's learning process and updating mastery estimates based on individual student exercises, PFA instead considers the number of correct exercises (wins) and incorrect exercises (fails) by a student on a given KC, along with a KC-level intercept to account for the relative difficulty of a KC. While BKT models KCs as independent entities, PFA can easily predict future performance while accounting for student knowledge of multiple KCs. More recent evaluations of PFA found it to be competitive with more contemporary KT models in certain scenarios \cite{gervet_when_2020,chu_predictiveness_2023}.

\paragraph{Deep Knowledge Tracing} \label{dkt}

Piech et al. \cite{piech_deep_2015} represents the first application of deep learning methods to the problem of KT. Broadly speaking, Deep Knowledge Tracing (DKT) is the application of a recurrent neural network (RNN) to sequences of exercise-response pairs to predict a student's ability to correctly answer future exercises. Due to its depth, the exact mechanisms by which DKT models student knowledge are less clear than with BKT or PFA. Though deep learning methods can provide performance gains over classical, "shallower" methods, Khajah et al. \cite{khajah_how_2016} achieve similar performance to the original DKT paper by analyzing certain advantages DKT has over BKT and extending BKT while keeping the underlying model and assumptions the same. In turn, later papers improve on DKT's learning gains by incorporating rich side information into the model \cite{zhang_incorporating_2017,wang_deep_2019}. 

\begin{figure}
    \centering
    \includegraphics[width=0.45\textwidth]{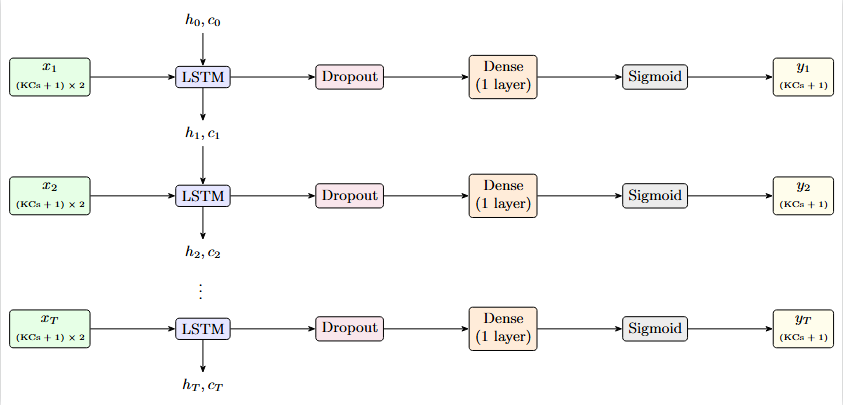}
    \Description[A block diagram of layers in a DKT Model]{ Shows the flow of data. Data flows through four layers 1-Hot-Encoded KC, LSTM, 1-Layer Dense, and the Output layer }
    \caption{Model architecture for DKT}
    \label{fig:DKT}
\end{figure}

\paragraph{Self-Attentive Knowledge Tracing} \label{sakt}

With the introduction of attention mechanisms to deep learning in Vaswani et al. \cite{vaswani_attention_2017}, time series classification problems across numerous domains achieved new state-of-the-art methods. KT was no exception to this, with Pandey \& Karypis \cite{pandey_self-attentive_2019} proposing an attention mechanism for knowledge tracing. Their aptly named Self-Attentive Knowledge Tracing (SAKT) surpassed previous models in performance, while simultaneously showing great promise as a more interpretable deep model, given the ability to visualize attention weights to understand particular exercises which the model weighted as more important or explanatory.

\begin{figure}[h]
    \centering
    \Description[A block diagram of Self-Attentive Knowledge Tracing Architecture]{Shows path of data from input to embed, multihead attention, LayerNorm, residual connection, and feedforward net with CE Loss}
    \includegraphics[width=0.3\textwidth]{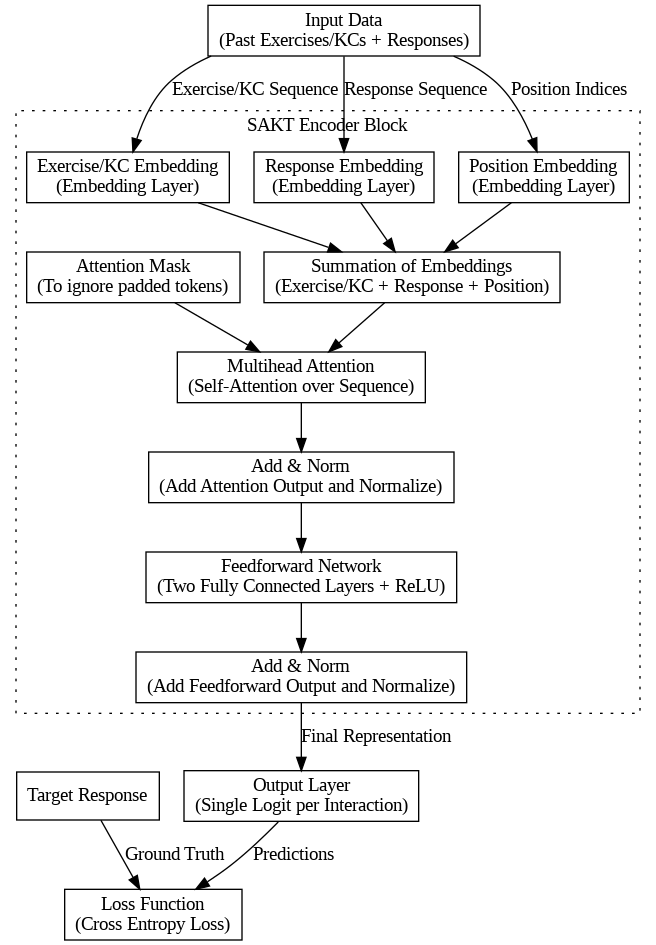}
    \caption{Model architecture for SAKT}
    \label{fig:SAKT}
\end{figure}

\subsection{Distributional Shifts} \label{dist-shifts}

Broadly speaking, for a given supervised learning problem with training set $X$ and labels $Y$, we are interested in modeling the joint probability distribution:
\begin{displaymath}
    P(X, Y)= P(Y|X)P(X)=P(X|Y)P(Y)
\end{displaymath}
Changes in this joint distribution can be categorized based on the part of this distribution that changes \cite{dmls_huyen_2022_ch8}:
\begin{enumerate}
    \item \textbf{Covariate Shift}, a change $P(X)$ while $P(Y|X)$ remains the same.
    \item \textbf{Label Shift}, a change in $P(Y)$ while $P(X|Y)$ remains the same.
    \item \textbf{Concept Drift}, a change in $P(Y|X)$ while $P(X)$ remains the same.
\end{enumerate}

Concept drift is perhaps the most difficult of these three shifts to adapt to, since a supervised learning model explicitly attempts to estimate $P(Y|X)$. Prior works have created methods to detect \cite{klinkenberg_detecting_nodate}, explain \cite{wang_explaining_2019}, and adapt to \cite{madireddy_adaptive_2019} concept drift. More recently, researchers have investigated how concept drift can impact common educational data mining (EDM) and learning analytics (LA) models. Levin et al. \cite{levin_evaluating_2022} explores how concept drift affects a variety of gaming detectors, finding that contemporary gaming detectors had more trouble generalizing to newer data than classic decision tree based methods, while Deho et al. \cite{deho_when_2024} found that concept drift in LA models is linked to algorithmic bias. These works highlight two distinct ways of attempting to quantify concept drift: through longitudinal model evaluation and through the application of concept drift detectors to log data.

\subsection{Prior Exploration of KT Generalizability} \label{prior}

Covariate Shift, Label Shift, and Concept Drift all have the potential to decrease the performance of KT models. Covariate and/or label shift could be introduced by an influx of new students, or a new curriculum being added to an OLP. Since KT models explicitly try to model the acquisition of knowledge, identifying when a model is susceptible to concept drift simultaneously raises questions about the underlying learning process.

This paper is not the first published work investigating the impact of changing student populations on knowledge tracing models. Lee et al. \cite{lee_knowledge_2023} investigated the stability of BKT model predictions over time and found that, while BKT is generally stable year-over-year, large, sudden shifts in student populations can have deleterious effects on model robustness. We wish to replicate and extend these findings by investigating the performance of other well-known KT models, including BKT, when applied to student interaction data spanning a longer time frame.

\section{Methods} \label{methods}
\begin{table*}[h]
\centering
\Description[A table showing dataset statistics]{ Shows Rows, Logs, Students, Unique KCs, and Percent Correctness by Academic Year from 2019 to 2024. }
\begin{tabular}{c|ccccc}
\multicolumn{1}{c|}{AY} & Total Rows & Assignment Logs & Unique Students & Unique KCs & \% Correct \\ \hline
2019\textendash2020               & 17,962,663 & 1,645,060       & 228,207         & 408        & 0.728              \\
2020\textendash2021               & 69,760,692 & 5,478,914       & 437,500         & 411        & 0.697              \\
2021\textendash2022               & 11,421,033 & 1,309,773       & 122,397         & 412        & 0.686              \\
2022\textendash2023               & 5,382,200  & 754,299         & 71,284          & 407        & 0.668              \\
2023\textendash2024               & 3,254,928  & 519,700         & 50,896          & 408        & 0.660             
\end{tabular}
\caption{Dataset sizes after filtering out ineligible problem logs.}
\label{tab:dataset-sizes}
\end{table*}

\subsection{Data Collection \& Preparation} \label{dataprep}
Data for this study was collected using the XXX OLP, spanning the five academic years between 2019\textendash2020 and 2023\textendash2024. Data not suitable for conducting Knowledge Tracing was filtered out, consisting of all data collected in the months of June, July, and August, as well as problem logs for non-computer-gradable questions, and all problem logs from problem set assigned fewer than 100 times total during the five academic years of interest. Summer student populations often differ greatly to the population of students using an OLP during the school year, while non-computer-gradable problems are incompatible with standard KT models, and removing low-use problem sets from the data lowers the likelihood of models differing solely due to out-of-vocabulary KCs and exercises. Information about the size of the data gathered from each academic year can be found in table \ref{tab:dataset-sizes}. The relative size of each year's data is worthy of note. Different years have great differences in the number of available logs, with the largest year having over twenty-one times the amount of total problem logs. Since the amount of available training data has a large impact on model fitness, this disparity in dataset sizes presents an issue.

To mitigate the impact of our dataset sizes, rather than using all available data for each year, we draw random samples from each available academic year. Randomly sampling \textit{user/exercise interactions} would isolate those rows from their surrounding context, while sampling \textit{per user} reintroduces concerns over differences in training set sizes, as the total number of exercises completed per user varies widely. Instead, we randomly sample 50,000 assignment logs, which are instances of a single student completing an assigned problem set. This allows us to draw samples of consistent size, since problem set length is more consistent, while collecting coherent sequences of student/exercise interactions in their full context. Our final dataset consists of ten such samples\footnote{In this paper, "sample" refers specifically to one of these random samples of 50,000 student/assignment interactions, \textit{not} individual examples of model inputs \& outputs.} per academic year, with samples containing 50,000 assignment logs each\footnote{These samples are available \href{https://osf.io/hvfn9/?view_only=b936c63dfdae4b0b987a2f0d4038f72a}{here}}.


\subsection{Study Design} \label{design}

In order to effectively investigate the susceptibility of KT models to concept drift, we need to establish baseline performance for each model on each target year and somehow evaluate models in a cross-year context. To measure within-year performance, we conducted a ten-fold cross validation, training one model per sample and evaluating it on the other nine samples\footnote{Rather than doing a classic 90/10 train-test split for crossvalidation, we opt to train on one sample and evaluate on the other nine to make sure all models are trained on roughly the same amount of data}. To investigate model performance across years, for each sample of a target year, we trained a model on the full sample and evaluated the fit model on one sample from all \textit{subsequent} years. While it's clearly possible to evaluate a model using data gathered \textit{before} the training year, doing so is more of an analytical tool, as in real systems possibly affected by concept drift, model accuracy decreases due to the introduction of \textit{later} data. Thus, we only evaluate models using data from their training year or later. Additionally, to explicitly investigate the effect of overparameterization on model performance over time, two different versions of SAKT were evaluated: one using KCs as model input and one using exercises directly.

\subsection{Model Implementations}

Each model was implemented in Python 3.12\footnote{These implementations, along with analysis code, are available \href{https://anonymous.4open.science/r/LongitudinalKnowledgeTracing24-4832/}{here}}, with the following differences. BKT was implemented with the forgetting parameter enabled via the hmmlearn package. After fitting models for each available KC in the training set, learned parameters were averaged to make a "best guess" KT model in the case of evaluating KCs that were not present in the training set. 
PFA was implemented using scikit-learn \cite{scikit-learn}, fitting separate covariates for wins and fails for each KC, along with a KC level intercept and parameters for KCs not present in the training set. Both SAKT and DKT were implemented in pytorch \cite{10.5555/3454287.3455008} and trained on NVIDIA A100 GPUs. Visual representations of model architectures can be found in figures \ref{fig:BKT}\textendash\ref{fig:SAKT}. For a full mathematical explanation of each model, please refer to appendix \ref{appendix}.

\subsubsection{Hyperparameter Tuning}
To tune the hyperparameters of our deep models, we used Bayesian hyperparameter optimization. Hyperparameters tuned for each model can be found in table \ref{tab:hyperparams}. Rather than tuning these hyperparameters with the samples constructed for our main study, we constructed a new validation sample by sampling 100,000 unique assignment logs from all available academic years. Using the optuna package \cite{optuna_2019}, we performed a four-fold cross-validation using the validation sample for each sampled hyperparameter combination, terminating after fifty rounds of cross-validation were performed.

\begin{table*}[h]
\centering
\Description[Table showing Parameters and their ranges for model types]{Shows optimal values for hyperparameters num_steps, batch size, d_model, num_epoch, dropout rate, learn rate, reg_lambda, num_heads \& learn_decay_rate}
\begin{tabular}{l|cccc}
Parameter        & Range                                                                  & DKT    & SAKT-E & SAKT-KC \\ \hline
\texttt{num\_steps}        & [20,100]                                                               & 40     & 60     & 100     \\
\texttt{batch\_size}       & [16,64]                                                                & 16     & 64     & 48      \\
\texttt{d\_model}          & [64,512]                                                               & 96     & 352    & 128     \\
\texttt{num\_epochs}       & \begin{tabular}[c]{@{}l@{}}DKT [100, 300]\\ SAKT [10, 40]\end{tabular} & 100    & 23     & 25      \\
\texttt{dropout\_rate}     & [0.1, 0.5]                                                             & 0.278  & 0.47   & 0.188   \\
\texttt{learn\_rate}       & [\num{1e-4}, \num{1e-2}]                                                           & \num{2e-3}   & \num{1e-4}   & \num{1e-4}    \\
\texttt{reg\_lambda}       & [\num{1e-6}, \num{1e-2}]                                                           & \num{1.4e-5} & n/a    & n/a     \\
\texttt{num\_heads}        & [2,4,8,16,32]                                                          & n/a    & 8      & 16      \\
\texttt{learn\_decay\_rate} & [0.7, 0.99]                                                            & n/a    & 0.7    & 0.868  
\end{tabular}
\caption{Learned Hyperparameters for DKT, SAKT-E, and SAKT-KC}
\label{tab:hyperparams}
\end{table*}

\section{Results} \label{results}

\begin{figure}[h]
    \centering
    \includegraphics[width=0.5\textwidth]{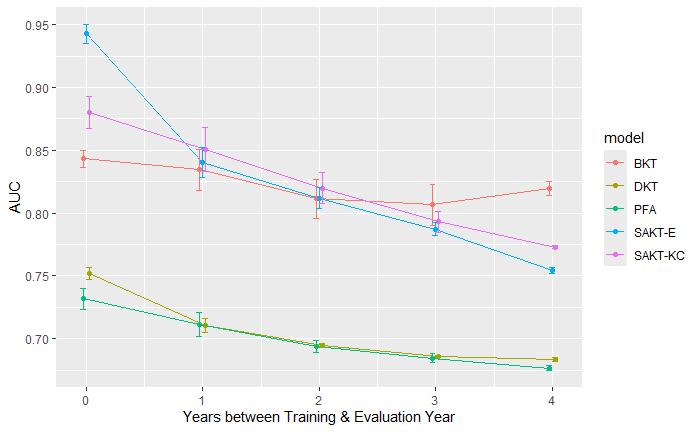}
    \Description[A Line Graph of mean AUC vs. training data age]{A Line with AUC values from 0.65 to 0.97 of 5 models. SAKT-E drops most from 0.95 to 0.77. PFA \& DKT change the least.}
    
    \caption{Mean AUC measurements vs. training data age. Error bars represent a 95\% CI for given training data age.}
    \label{fig:model-eval-auc}
\end{figure}

Model evaluation results can be found in figure \ref{fig:model-eval-auc} which shows the AUC of each model compared to the years since the model was trained organized by model. Every model tested by our method had decreasing AUC over time. All five models have significant correlations between model fit metrics and the number of years between training \& evaluation data (see table \ref{tab:spearmans} for estimated correlation coefficients) indicating that increased age since training worsens model performance. BKT is notable for this relationship not being strictly monotonic. This suggests that after 3 or 4 years since training, the decrease in performance may have hit a maximum and they may continue to perform similarly while the other models will take more continue to have decreasing performance past 4 years. BKT and PFA lost the least predictive power across time, losing only $\approx0.03$ \& $\approx0.05$ AUC, while SAKT-E was the most impacted, losing $\approx0.17$ AUC.

\begin{figure}[h]
    \centering
    \includegraphics[width=0.5\textwidth]{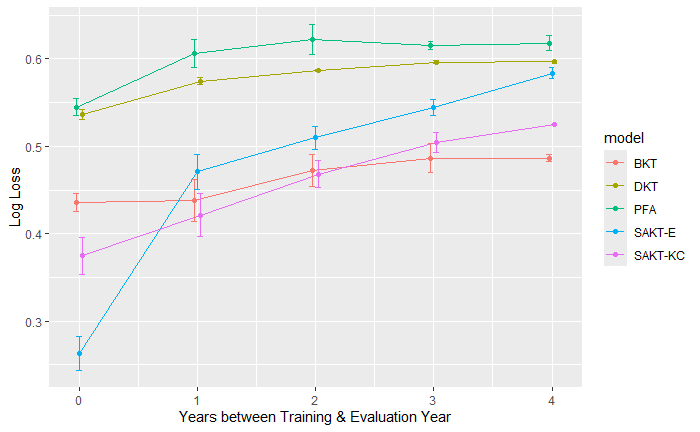}
    \Description[A Line Graph of Log Loss lines for 5 different models]{5-line graph with values from 0.25 to 0.65, centered around 0.5. SAKT-E has the greatest change, starting at 0.25, and jumping to 0.56}
    \caption{Mean Log Loss measurements vs. training data age. Error bars represent a 95\% CI for given training data age.}
    \label{fig:model-eval-logloss}
\end{figure}

Figure \ref{fig:model-eval-logloss} shows the log loss of each model compared to the number of years since the models were trained. All models had increasing log loss as the number of years increased, with SAKT-E having the greatest increase of $\approx0.3$ and DKT the smallest of $\approx0.05$. SAKT-E and SAKT-KT both continue to decrease in performance across all years, while PFA, DKT and BKT stop worsening around the final year. This continues to suggest that BKT, DKT and possibly PFA, which had the smallest drop in performance for the AUC model, may have reached their worst performance and continue to have a steady log loss. The final figure of F1 Score and years since training shows similar results, with SAKT-E and SAKT-KC decreasing the most and BKT DKT and PFA decreasing less, and leveling out towards the end.

\begin{figure}[h]
    \centering  
    \includegraphics[width=0.5\textwidth]{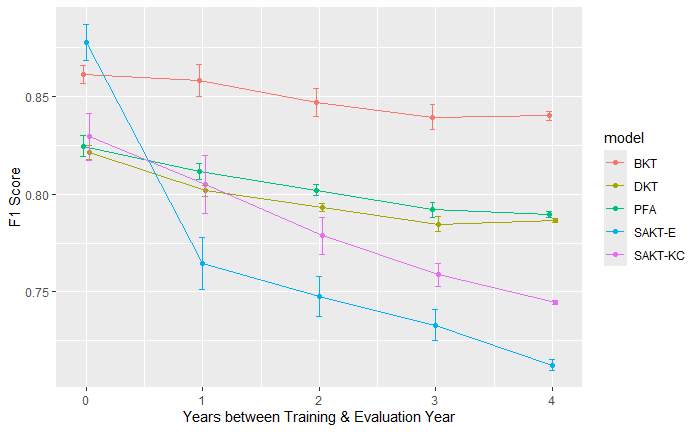}
    \Description[Trajectory of all 5 model type F1 scores]{}
    \caption{Mean F1 score measurements vs. training data age. Error bars represent a 95\% CI for given training data age.}
    \label{fig:model-eval-f1}
\end{figure}

Figure \ref{fig:model-eval-f1} shows the F1 score of each model compared to the number of years since the models were trained. Unlike with AUC, \textit{every} model's performance was strictly decreasing. Similar to the trends in AUC and Log Loss, the two SAKT models  showed the sharpest declines, with SAKT-KC's decline looking broadly linear. SAKT-E experiences a sharper decrease between zero and one years between training \& evaluation, then appears to decline at a similar rate to SAKT-KC. 

Finally, to compare the rates of model performance loss, we performed the following fixed effects regression on all three reported metrics:
\begin{displaymath}
    \text{metric}=\alpha_m + \beta y + \beta_{m}y
\end{displaymath}
Where $\alpha_m$ is the estimated fixed effect for model $m$, $y$ is the number of years between training and evaluation data, $\beta$ is our estimated rate of performance loss for the reference category and $\beta_m$ is the interaction term between $\alpha_m$ and $\beta$. Results from these three regressions can be found in tables \ref{tab:reg-results-auc}\textendash\ref{tab:reg-results-f1}.


\begin{table}[h]
\centering
\Description[Table showing architecture of each model]{Shows proposed architecture, Logical Unit, and number of tunable parameters for BKT, PFA, DKT, and both SAKT variants}
\begin{tabular}{c|cccc}
Model  & Proposed & Logical Unit  & Tunable Params \\ \hline
BKT    & \cite{corbett_knowledge_1994} & KC & 2,012\\
PFA     & \cite{pavlik_performance_2009} & KC & 1,208\\
DKT     & \cite{piech_deep_2015} & KC  &734,026\\
SAKT-KC & \cite{pandey_self-attentive_2019} & KC & 262,982\\
SAKT-E  & \cite{pandey_self-attentive_2019} & Exercise & 12,400,658                              
\end{tabular}
\caption{Number of trainable parameters of each model.}
\label{tab:num-params}
\end{table}

\begin{table}[h]
\centering
\Description[Table with estimated Spearman Correlations for all reported Evaluation Metrics]{}
\begin{tabular}{c|ccc}
Model   & AUC      & Logistic Loss & F1 Score \\ \hline
BKT     & $-0.374$ & $0.411$       & $-0.428$ \\
PFA     & $-0.716$ & $0.779$       & $-0.721$ \\
DKT     & $-0.917$ & $0.918$       & $-0.861$ \\
SAKT-E  & $-0.883$ & $0.872$       & $-0.810$ \\
SAKT-KC & $-0.695$ & $0.693$       & $-0.673$
\end{tabular}
\caption{Estimated Spearman's $\bm{\rho}$ for each model fit statistic. All estimated coefficients have $\bm{p}$ \textbf{\textless{}} $\bm{0.001}$.}
\label{tab:spearmans}
\end{table}

\begin{table}[h]
\begin{tabular}{cccc}
Coefficients                  & Estimate       & Std. Err      & p-value        \\ \hline
Years Between (YB)               & $-0.010$       & \num{2.12e-3} & \num{1.63e-5}  \\
BKT                     & $0.842$        & \num{3.87e-3} & \\
PFA                     & $0.729$       & \num{3.87e-3}&  \\
DKT                     & $0.743$       & \num{3.87e-3} & \\
SAKT-E                  & $0.921$        & \num{3.87e-3} &  \\
SAKT-KC                 & $0.879$        & \num{3.87e-3} &  \\
YB$\times$PFA     & \num{-4.95e-3} & \num{3.00e-3} & 0.496        \\
YB$\times$DKT     & \num{-9.70e-3} & \num{3.00e-3} & \num{7.58e-3}  \\
YB$\times$SAKT-E  & $-0.039$       & \num{3.00e-3} & \num{1.01e-33} \\
YB$\times$SAKT-KC & $-0.018$       & \num{3.00e-3} & \num{2.94e-8} 
\end{tabular}
\caption{Regression results for AUC (adjusted $\mathbf{R}^2=0.836$), with BKT as reference category}
\label{tab:reg-results-auc}
\end{table}

\begin{table}[h]
\begin{tabular}{cccc}
Coefficients               & Estimate      & Std. Err      & p-value         \\ \hline
Years Between (YB)                & 0.016         & \num{3.51e-3} & \num{5.32e-5}   \\
BKT                  & 0.433         & \num{6.41e-3} & \\
PFA                  & 0.562         & \num{6.41e-3} & \\
DKT                  & 0.545         & \num{6.41e-3} & \\
SAKT-E               & 0.312      & \num{6.41e-3} &    \\
SAKT-KC              & 0.378      & \num{6.41e-3} &    \\
YB$\times$PFA     & \num{5.70e-3} & \num{4.96e-3} & 0.597           \\
YB$\times$DKT     & \num{1.62e-3} & \num{4.96e-3} & 0.745           \\
YB$\times$SAKT-E  & 0.070         & \num{4.96e-3} & \num{4.75e-39}  \\
YB$\times$SAKT-KC & 0.025         & \num{4.96e-3} & \num{5.21e-6}  
\end{tabular}
\caption{Regression results for Log Loss (adjusted $\mathbf{R}^2=0.737$), with BKT as reference category}
\label{tab:reg-results-ll}
\end{table}

\begin{table}[h]
\begin{tabular}{cccc}
Coefficients               & Estimate       & Std. Err      & p-value        \\ \hline
Years Between (YB)         & \num{-6.55e-3} & \num{1.82e-3} & \num{2.33e-3}  \\
BKT                  & 0.862          & \num{3.32e-3} &           \\
PFA                  & 0.823       & \num{4.69e-3} &  \\
DKT                  & 0.817       & \num{4.69e-3} &  \\
SAKT-E               & 0.850       & \num{4.69e-3} &  \\
SAKT-KC              & 0.828       & \num{4.69e-3} &  \\
YB$\times$PFA     & \num{-3.31e-3} & \num{2.57e-3} & 0.597          \\
YB$\times$DKT     & \num{-3.96e-3} & \num{2.57e-3} & 0.496          \\
YB$\times$SAKT-E  & $-0.038$       & \num{2.57e-3} & \num{9.50e-42} \\
YB$\times$SAKT-KC & $-0.016$       & \num{2.57e-3} & \num{6.13e-9} 
\end{tabular}
\caption{Regression results for F1 score (adjusted $\mathbf{R}^2=0.632$), with BKT as reference category}
\label{tab:reg-results-f1}
\end{table}

\section{Discussion} \label{discussion}

Our results suggest that all KT models we tested are vulnerable to concept drift under some conditions. This includes BKT, though BKT seems the most reliably robust by all three recorded metrics. Every other model we assessed had more extreme rates of performance loss (see the interaction terms in tables \ref{tab:reg-results-auc}\textendash\ref{tab:reg-results-f1}). It is also telling that model performance degradation seems linked to models being used outside of their temporal context. That is, model degradation is more pronounced when evaluated on data far newer than the data used to train the original model. When evaluated on data from four years later, every model examined lost at \textit{least} 0.05 AUC. However, for BKT, DKT, and PFA the performance degradation appears to slow down if not completely halt after a certain time frame. This suggests that there may be a certain number of years after which the maximum model degradation has occurred and model's retain their predictive power year after year. For BKT which had a final AUC of 0.82 this would indicate that it may be viable to use well past the year of its training data. Similarly, DKT and PFA may continue to be used past the years on which they were trained without expecting serious amounts of model degradation, with the caveat that they did not perform as well as other any other models to begin with. Finally, SAKT-E and SAKT-KC both have significant predictive power the first 1-2 years after they are trained, however appear to be degrading with no signs of stopping suggesting they suffer the most from concept drift. 

Based on our findings, KT model complexity (as measured by the number of trainable parameters) may not be directly linked to lower model robustness as theorized in Levin et al. \cite{levin_evaluating_2022}. While SAKT-E exhibited the steepest drop-off in model performance over time, particularly in the year immediately following its training, SAKT-KC also experienced statistically significant losses in AUC despite having orders of magnitude fewer tunable parameters. DKT which had more parameters than SAKT-KC but fewer than SAKT-E, and was far more robust than both. BKT and PFA, which had the fewest tunable parameters by orders of magnitude, experienced a noticeable but minimal degradation. Rather, the main factor which contributes to model degradation appears to be the presence of the attention mechanism in both SAKT-KC and SAKT-E. SAKT-E is the top performing model and SAKT-KC is the next-best model by AUC the when evaluated on the year in which they were trained, however, both exhibit steep decline as the years between evaluation and training increase.

Given that the two SAKT models had significantly higher rates of performance decline than the other three models, it seems that the attention mechanism of SAKT may cause its downfall. One advantage of self-attentive models in other domains is their ability to capture long-range dependencies. This could serve as a detriment since students' performance is more correlated to their performance on nearby problems rather than how they did on problems two months ago. Further, this would explain, in part, why across years the SAKT models perform significantly worse, as longer range patterns derived from one year lose their explanatory power in future years due to curriculum changes, ordering of skills taught, and other sources of concept shift. Such close-range dependencies would favor BKT and DKT due to their ability to estimate ability on certain skills and how students have performed on recent problems.


\subsection{Limitations \& Future Work} \label{limits}

While our findings broadly suggest that KT models are susceptible to concept drift, there are notable limitations in our analysis. Our model-focused approach to measuring concept drift cannot describe \textit{how} the distribution of student responses has changed, nor explain factors which could be causing said change. Clearly \textit{something} about the interactions students have within an OLP changes through time, and future works could employ a data-centric approach to detecting concept drift alongside evaluating models through time. Alternatively, a more thorough investigation of the learned attention weights of the different SAKT models could yield more insight, either identifying what parts of the learning process change over time or verifying that the model is more sensitive to noisy sequences. We limited our analysis to basic implementations of our four KT models. As discussed in Shen et al. \cite{shen_survey_2024}, our four choices of models represent broader "families" of KT models, and novel KT modeling is an area of active research in EDM. Exploring how extensions to these model families impacts robustness would also give more insight into the relationship between model complexity and generalizability.

\section{Conclusion} \label{conclusion}

Expanding upon previous findings, this study indicates that many popular families of Knowledge Tracing models can lose predictive power over time. BKT and PFA exhibit the slowest rates of decline, while more recently proposed attention-based models decline faster. These findings indicate that the underlying process of student learning (as monitored through student interaction logs) may not be as stable as previously theorized. 

These results have multiple implications for researchers and practitioners in the EDM field. There is a long-standing discussion surrounding the pros and cons of deep learning models in an educational context. The choice between "simpler"/more explainable models and more complex but less explainable models involves considerations of data availability, the desire for explainable models, and ease of model fitting, among other concerns. Based on our findings, we posit that model longevity should be another consideration when deciding how to model student learning. If a model is needed only for the short-medium term, a deeper model may well be more appropriate. In situations where a model may need to be valid over a longer duration, our results indicate a simpler model may be a better choice. Deeper models may indeed be more performant when applied to a data-abundant environment, and when refit every two-to-three, but in more data-sparse contexts, simpler models may be more advisable. We encourage other researchers interested in proposing new KT models to explore their models' fitness over longer time frames rather than relying on a single benchmark dataset, and in the interest of pursuing this goal, we are releasing the samples described in section \ref{dataprep} with this paper.

Investigations targeting student interaction data could yield further insights into how and why student behavior changes, and may be key in creating models and training schedules that are robust in the presence of concept drift. There are also many more common educational models, such as affect detectors and psychometric models, which may have differing levels of robustness to changing student populations. Understanding which models lose accuracy over time and why is an essential step in understanding how student learning behavior changes over time.

\section*{Acknowledgements}
Redacted for blinding purposes.


\bibliographystyle{abbrv}
\bibliography{sigproc}

\appendix
\section{Model Explanations}
\label{appendix}
\subsection{BKT}
\begin{displaymath} \label{eq:bkt1}
    P(L_1) = P(L_0) 
\end{displaymath}
\begin{displaymath} \label{eq:bkt2}
    P(L_t|C_t = 1) = \frac{P(L_t)\cdot(1-P(S))}{P(L_t)\cdot(1-P(S))+(1-P(L_t))\cdot P(G)}
\end{displaymath}
\begin{displaymath} \label{eq:bkt3}
    P(L_t|C_t = 0) = \frac{P(L_t)\cdot P(S)}{P(L_t)\cdot P(S)+(1-P(L_t))\cdot(1-P(G))}
\end{displaymath}
\begin{displaymath} \label{eq:bkt4}
    P(L_{t+1} = P(L_t|C_t)\cdot (1-P(F)) + (1-P(L_t|C_t))\cdot P(T)
\end{displaymath}
\begin{displaymath} \label{eq:bkt5}
    P(C_{t+1}) = P(L_{t+1})\cdot(1-P(S))+(1-P(L_{t+1}))\cdot P(G)
\end{displaymath}

$P(L_t)$ is the probability that the student has mastered the relevant KC at time step $t$, $P(C_t)$ is the probability of said student getting the next exercise of that KC correct. $P(L_0)$, $P(G)$, $P(S)$, $P(T)$, and $P(F)$ are the probabilities for prior knowledge, guess, slip, transfer (learning), and forgetting, respectively, and are learned via expectation maximization \cite{corbett_knowledge_1994}.

\subsection{PFA}

\begin{displaymath}
    m(i,j\in KCs, s, f) = \sum_{j \in KCs}{\beta_j +\gamma_j s_{i,j} + \rho_j f_{i,j}}
\end{displaymath}
\begin{displaymath}
    P(m) = \frac{1}{1 + e^{-m}} = \sigma(m)
\end{displaymath}

In this model, $i$ represents a student, while $j$ represents a KC. $s_{i,j}$ and $f_{i,j}$ represent collective successes and failures of student $i$ on KC $j$ \cite{pavlik_performance_2009}. 

\subsection{DKT}
Our implementation of DKT uses the standard LSTM \cite{10.1162/neco.1997.9.8.1735} fed into a dense output layer. An LSTM computes the following function:
\begin{displaymath}
    i_t = \sigma(W_{ii}x_t+b_{ii}+W_{hi}h_{t-1}+b_{hi})
\end{displaymath}
\begin{displaymath}
    f_t = \sigma(W_{if}x_t+b_{if}+W_{hf}h_{t-1}+b_{hf})
\end{displaymath}
\begin{displaymath}
    g_t=\tanh (W_{ig}x_t+b_{ig}+W_{hg}h_{t-1}+b_{hg})
\end{displaymath}
\begin{displaymath}
    o_t = \sigma((W_{io}x_t+b_{io}+W_{ho}h_{t-1}+b_{ho})
\end{displaymath}
\begin{displaymath}
    c_t = f_t \odot c_{t-1}+i_t\odot g_t
\end{displaymath}
\begin{displaymath}
    h_t = o_t \odot \tanh(c_t)
\end{displaymath}

where $i_t$, $f_t$, $g_t$, $o_t$, $c_t$, $h_t$ are the input, forget, output, cell, and hidden state at time $t$, $x_t$ is the model input at time $t$, $\sigma$ is the sigmoid function and $\odot$ is the element-wise product. All instances of $W$ and $b$ are trainable parameters.

\subsection{SAKT}
Our implementation of SAKT uses architecture initially described in Pandey et al's SAKT paper \cite{pandey_self-attentive_2019}. The architecture using exercises as features is fittingly called SAKT-E. An added change from the original paper is the addition of KCs as a potential feature for embedding, for which the architecture is appropriately titled SAKT-KC.  
\begin{table}[h!]
\centering
\Description[Table with Notation for SAKT Architecture]{Table outlining list of notations for mathematical shorthand employed in explanation of SAKT architecture}
\begin{tabular}{|c|c|}
\hline
\textbf{Notations} & \textbf{Descriptions} \\
\hline
$B$      & Batch size \\
$T$      & Sequence length \\
$d$      & Embedding dimension \\
$K$      & Total number of exercise/KC IDs \\
$X$      & Past learner responses, $X \in \mathbb{R}^{B \times T \times 2}$ \\
$R$      & Past binary responses $R \in \{0, 1\}$ \\
$y$      & Ground truth labels ($y \in \{0, 1\}$) \\
$C$      & Binary ground truth labels ($C \in \{0, 1\}$) \\
$L$      & Number of unique exercises \\
$E_x$    & Trainable exercise/KC ID embedding matrix \\
$P_x$    & Trainable position embedding matrix \\
$R_x$    & Trainable Response embedding matrix \\
$W$ & Trainable Weight Matrix\\
\hline
\end{tabular}
\caption{Notations for SAKT Architecture}
\end{table}

\subsubsection{Embedding Layers}
\noindent
\begin{displaymath}
E_x = \{e_t \mid e_t = \hat{E}[E_t], \ E_t \in \{0, 1\}\}, \quad \hat{E} \in \mathbb{R}^{B \times T \times d}
\end{displaymath}
\begin{displaymath}
R_x = \{r_t \mid r_t = \hat{R}[R_t],\ R_t \in \{0, 1\}\}, \quad \hat{R} \in \mathbb{R}^{B \times 2 \times d}
\end{displaymath}
\begin{displaymath}
P_x = \{p_t \mid p_t = E_g[t],\ t \in [1, T]\}, \quad E_g \in \mathbb{R}^{ B \times T \times d}
\end{displaymath}
\begin{displaymath}
Z = E_x + R_x + P_x    
\end{displaymath}

\subsubsection{Self-Attention Mechanism}
\begin{displaymath}
Q = ZW^Q, \quad K = ZW^K, \quad V = ZW^V
\end{displaymath}
\begin{displaymath}
\text{Attention}(Q, K, V) = \text{softmax}\left(\frac{QK^\top}{\sqrt{d}}\right)V     
\end{displaymath}
\begin{displaymath}
\text{Head}_h = \text{Attention}(QW_h^Q, KW_h^K, VW_h^V)
\end{displaymath}
\begin{displaymath}
 \text{MultiHead}(Q, K, V) = \text{Concat}(\text{Head}_1, \dots, \text{Head}_H)W^O   
\end{displaymath}

Here, softmax refers to the softmax function, formally defined as $$Softmax(x_i) = \frac{e^{x_i}}{\sum \limits _{j=1} ^ n  e^{x_j}}$$ where n is the number of classes. 

\subsubsection{Residual Connections and LayerNorm}
\begin{displaymath}
O = \text{LayerNorm}(Z + \text{MultiHead}(Q, K, V))
\end{displaymath}
\begin{displaymath}
 F(O) = \max(0, OW_1 + b_1)W_2 + b_2   
\end{displaymath}
\begin{displaymath}
 O = \text{LayerNorm}(O + F(O))   
\end{displaymath}
\begin{displaymath}
 \hat{y}_t = \sigma(W_yO_t + b_y)   
\end{displaymath}

LayerNorm here refers to the function initially defined in Ba, et.al's paper \cite{ba2016layernormalization} on normalization of neurons, which reduces training time significantly. This formula is given by $y =  \frac{x-E[x]}{\sqrt{Var(x) + \epsilon}} $ where $\epsilon$ is a small value added to avoid division by 0. In the calculation of our binary prediction $\hat{y}$, $\sigma$ represents the sigmoid activation function, or $\frac{1}{1+e^-x}$ to squash the output into the range of (0, 1) for evalution with the loss function. 
\subsubsection{Loss Function}
\begin{displaymath}
\mathcal{L} = -\frac{1}{K} \sum_{i=1}^K \left[ y_i \log(\hat{y}_i) + (1 - y_i)\log(1 - \hat{y}_i) \right]    
\end{displaymath}
Here, we use the standard binary cross entropy loss, with K representing the number of observations. 
\balancecolumns
\end{document}